# FLEXIBLE LOG FILE PARSING USING HIDDEN MARKOV MODELS

Nadine Kuhnert and Andreas Maier

Pattern Recognition, Friedrich-Alexander University, Erlangen-Nuremberg, Germany
nadine.kuhnert@fau.de andreas.maier@fau.de

## ABSTRACT

*We aim to model unknown file processing. As the content of log files often evolves over time, we established a dynamic statistical model which learns and adapts processing and parsing rules. First, we limit the amount of unstructured text by focusing only on those frequent patterns which lead to the desired output table similar to Vaarandi [10]. Second, we transform the found frequent patterns and the output stating the parsed table into a Hidden Markov Model (HMM). We use this HMM as a specific, however, flexible representation of a pattern for log file processing. With changes in the raw log file distorting learned patterns, we aim the model to adapt automatically in order to maintain high quality output. After training our model on one system type, applying the model and the resulting parsing rule to a different system with slightly different log file patterns, we achieve an accuracy over 99%.*

## 1. INTRODUCTION

Predominantly with the goal of monitoring, almost any computer system produces log files containing information about procedures, events, issues, and errors. These log files are generated during operation mostly in the form of text or xml files. They add up to a huge amount of information ready to be interpreted. By parsing the log files, valuable information is extracted which can then be further processed into knowledge. With software updates and version changes, log file contents and patterns might change. E.g. some Key Performance Indicators (KPIs) are only logged since a specific version. Kuhnert et al. [1] tackled this issue of the body region being only logged by the latest Magnetic Resonance Imaging systems. They applied clustering methods in order to learn the examined body region from the scan parameters. Thus, they applied their learnt clustering algorithm to logged scan parameters from earlier software versions and could complete the examined body region information in the respective result tables. Furthermore, another issue of changes in the logged events is that rigidly implemented parsing rules will fail and lead to incomplete extracted data. Practice has shown that in some cases patterns are flexible enough, in other cases patterns are manually adjusted in time before patterns fail due to changed log file content. However, those two described scenarios do not always apply which leads to failing patterns and missing output data. As this is the very first step of turning raw data into actionable insights, it has to be considered a major problem for all further data analysis steps. Figure 1 depicts exactly this scenario, where two systems produce log files during operation. The systems differ in their software version. This can lead to varied content in the log file. We highlighted the difference in the log files in grey. Applying the same, rigid parser can result in complete tables for system A on the one hand, and incomplete tables for system B on the other.

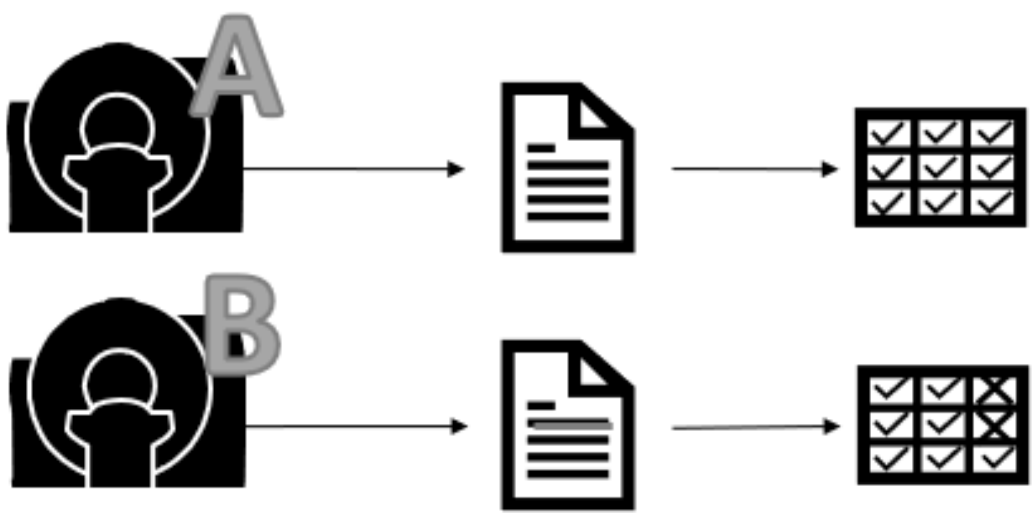

Figure 1. Problem Statement: Emerging log file entries processed by rigid parsers will lead to missing entries in output data.

Our goal is to build a parser which adapts automatically to gradual changes in the input data. The input data we apply our algorithms on has been produced by Computed Tomography (CT) systems during operation. Next to the medical imaging data, CTs constantly write events into text files which are subject to our research. As described in Maier et al. [2], during acquisition, a CT system applies X-rays from different angles and records multiple projection images. Thus, 3-D reconstruction enables cross-sectional views of the examined objects. Figure 2 shows an exemplary CT scanner. The examined body is exposed to X-rays which can harm healthy tissue. Patient's health and regulatory limitations require to measure the radiation dose being applied to the patient. The exposed radiation dose is one Key Performance Indicator (KPI) which is recorded in an event log file and is often denoted as Computed Tomography Dose Index (CTDI).

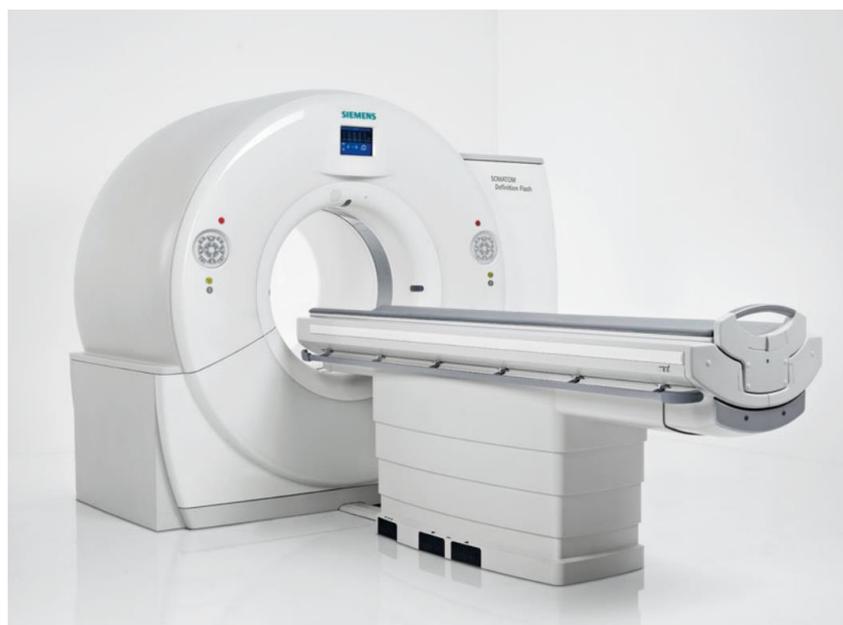

Figure 2. Computed Tomography Scanner. [2]

## 2. STATE OF THE ART

The high importance of turning plain log files into usable knowledge correlates with the large number of literature handling log file analysis. Already in 1993, Hansen and Atkins [3] applied algorithms for system monitoring and notification.

The most common methods of extracting information from log files base on detecting known fault types using regular expressions [4][5]. For more flexible parsing methods, several data mining approaches have been applied in order to discover trends and correlations without knowing the exact pattern a priori [6][7][8][9]. For example, Vaarandi [10] uses clustering algorithms with the goal of detecting frequent patterns as well as identifying anomalous log file lines.

Since log file entries are discrete, sequential data, applying Markov models is natural choice. The statistical concept of Hidden Markov Models (HMM) was already in 1966 originated by Baum and Petrie [11], whereas Rabiner [12] took that concept further into practice. A HMM is a statistical signal model with unobservable (hidden) states whose likelihood only depends on the preceding state (Markov property). Emissions, also called outputs, are observable states connected to the hidden states by emission probabilities. Thus, a HMM is fully described by a set of hidden states, emissions, starting probabilities, transition probabilities and emission probabilities. By setting up a HMM, three fundamental problems can be addressed. First, the evaluation problem can be addressed using the forward-backward algorithm. Thus, the question of what the probability of a particular output sequence given the model is can be answered. Secondly, given the model and a given output sequence, we want to find the most likely sequence of hidden states. This is solved by the Viterbi algorithm. Lastly, the so-called learning problem addresses finding the most likely set of state transition and emission probabilities given a set of emissions, which is solved by the Baum-Welch algorithm and used in fitting new data to a previously learnt HMM.

Furthermore, Yamanishi and Maruyama [9] tackle the issue of evolving sequences of events in the field of network failure monitoring and propose combining HMM mixture with adaptive learning of parameters to achieve dynamic modeling and adaptive tracking. More general term of data transformation for an increase of information content is data wrangling which is discussed by Endel et al. [13] in "how to make data useful again".

Yamanishi and Maruyama [9] focussed on analysing the content of log files and mining symbolic data. With our work, we contribute to research by using ideas from [9], applying an adjusted version of [10] and introducing the new concept of using HMMs for adaptive parsing.

## 3. MATERIALS AND METHODS

Our goal is to extract specific, reoccurring information out of the raw log file while the according entries and rules will emerge with software version updates. Here, we present our approach using the example of amount of dose, called "ctdi", which is applied during a Computed Tomography (CT) scan. We reach our goal by implementing a processing pipeline as illustrated in Figure 3.

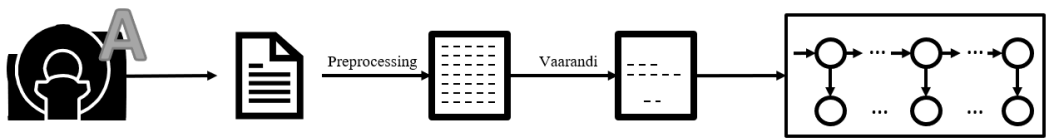

Figure 3. Processing pipeline of an adaptive parser using HMM.

First, we apply some text preprocessing steps such as tokenization, stemming and lower capitalization. Furthermore, we remove all English stopwords and punctuation and receive a cleared set of tokens.

Based on that preprocessed data set we implemented and applied a slightly modified version of Vaarandi's "data clustering algorithm for mining patterns from event logs" [10] in order to find common structures. Vaarandi's original approach consists of mainly three steps: Within the first run through the available data set frequent words are found by simply counting occurrences and applying a threshold. Based on that data summary in a second run cluster candidates are built. Cluster candidates contain one or more words that were detected to be frequent words in the first step as well as occur in the same line of text. Support values per cluster candidate are calculated which represent the number of lines containing those cluster candidates. In the third and last step the final clusters are selected from the cluster candidates by filtering the support values greater or equal to a certain threshold.

In our approach, we have to expect irregular order of our events because of software adaptions. Thus, we do not consider the word's position and order unlike suggested by Vaarandi. As we focus on the dose value in this paper, we can assume to find only one kind of KPI to be parsed per line and parsing pattern. Thus, we use the amount of KPI values (e.g. 20.0) as a threshold rather than setting or optimizing a specific threshold as proposed by Vaarandi. The result is a vector of clusters representing frequently occurring types of lines. We further reduce the resulting clusters to the number of expected distinct KPIs (e.g. "ctdi") and choose the most likely clusters.

This is followed by the set up and training of the HMM based on those data. The found cluster is interpreted as a parsing pattern and used as the states of our HMM. The starting probability $p_{si}$ reflects the frequency of the tokens, accordingly. A first-order Markov model depicts a sequence of states, where one state's likelihood can be predicted using its conditional probability given one preceding state. Therefore, next to the probability of the HMM's entry point, information about the subsequent state's likelihood $pt_{i,j}$ of occurrence is stored in a transition probability matrix by utilizing the concept of bigrams. A schematic presentation of the model is displayed in Figure 4. The initial starting probability is indicated by arrows drawn from *Start* to the respective $state_i$. Circles in the middle lane represent the different states. Arrows between the states illustrate likelihood of one state being followed by the other. Furthermore, dotted lines connect states to emissions and specify the emission probability, respectively.

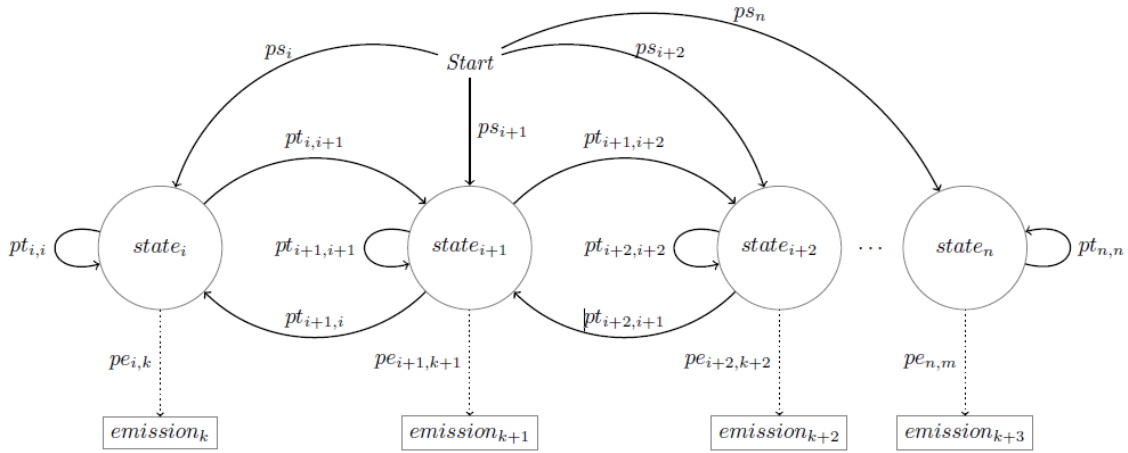

Figure 4. Parsing one element represented by a HMM.

Thus, our HMM is built and fully described by a vector of states, emissions, starting, transition and emission probability matrix. The states are hidden and represent entries in the original log file. Emissions can be observed and found as dose values in the output tables of the parsing process. $pe_{i,k}$ describes the probability that a certain emission k follows a given state i. This defines our HMM exhaustively.

Once we built the HMM representing a flexible version of a parsing pattern for dose values on one system type (Data *A*), we can apply the abstract pattern to data of a different system type (Data *B*) in order to extract the relevant dose information. Furthermore, we find the most probable states for a new observation sequence using Viterbi [12]. This is illustrated in Figure 5, as well as using Baum-Welch [12] to fit our model to new data. This implies that we update the transition and emission probabilities which connect the states with each other and states with emissions, accordingly.

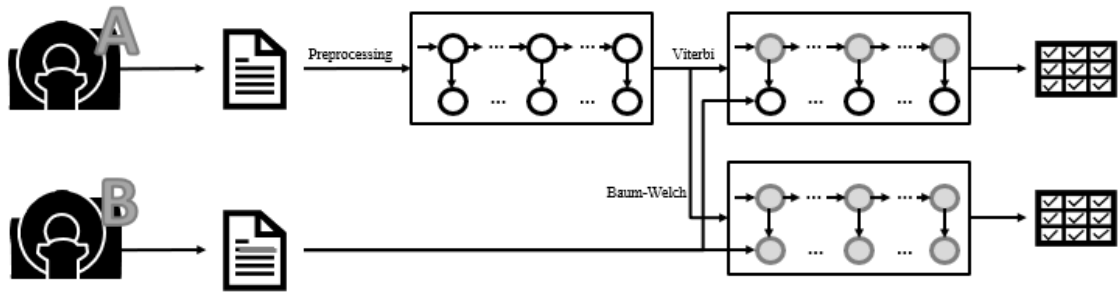

Figure 5. Pipeline of training the HMM on Data A and applying the implicit pattern on Data B. The HMM representing a flexible version of the parsing pattern is adapted to new Data B in two ways, using Viterbi as well as Baum-Welch, in order to get the correctly parsed output table.

## 4. RESULTS

We trained and tested our new, flexible parsing approach on two data sets as shown in Figure 4. The input data constitutes log file entries and accordingly parsed values of applied dose from one CT system over the period of December 2018, further referenced by data set *A*. For a first evaluation we split *A* into training and testing using a stratified split. In the following for the purpose of testing the trained algorithm on the very same system, we will refer to 70% of data set *A* by $A_{train}$ and use the term $A_{test}$ for the remainig 30% of *A*. Henceforth, we will call data from a different CT system of type B from December 2018 for validation purposes of our algorithm's adaptability as data set $B_{valid}$, accordingly. This is illustrated in Figure 6.

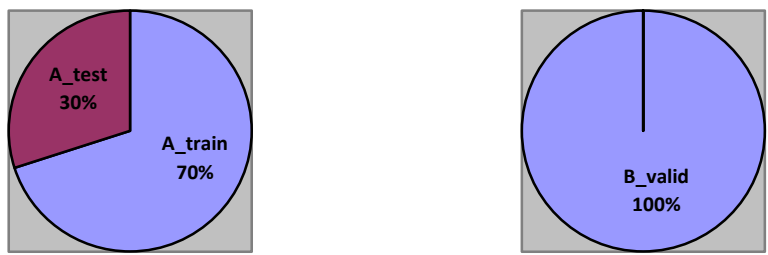

Figure 6. Two data sets from different CT system types are used, whereas the data of system A is split into training and testing. Validation is performed on data of system B.

### 4.1. Intermediate Results along the Pipeline

Analogously to our processing pipeline, we present intermediate results and evaluate all steps in order to assess their importance. In Figure 7, we present an anonymized example event text of a raw log file which carries information about a specific scan. Among that information, also the amount of applied dose can be found. This event text is tied to a time stamp, event type and event ID.

| &Load scan protocol&,@Patient LOID@=#2.0.123456#,@Scan@=#1#,@ScanUID@=#1.3.12.2.1107.5.1.4.83004.1234567890#,@Scan protocol name@=#rot00#,@Organ characteristics@=#MlOrgCharAbdomen#,@Body size original@=#MlAdult#,@Scan entry name@=#rot00#,@Kind@=#MlRot#,@Entry Mode@=#standard#,@AutoRange@=#Cont#,@kV@=#120#,@mAs@=#250#,@CARE Dose@=#Off#,@AEC@=#Off#,@CTDI@=#16.660#,@DLP@=#59.975#,@Slice@=#0.6#,@Scan start@=#MlRangeStartAuto#,@Slice Width Collimated@=#60#,@No Of Acquisition Slices@=#60#,@CBC@=#Off#,@Scan trigger@=#MlScanTriggerAuto#,@No of scans@=#1#,@Examination time@=#0.500000#,@ScanTime@=#1.000#,@RotTime@=#0.500#,@RotKind@=#Normal#,@CurrentPeak@=#250#,@DoseModulationType@=#MlNoModulation#,@Focus@=#MlSmallFocus#,@Anodespeed A@=#120#,@StartDelay@=#2.000#,@NoOfClustersPerRange@=#1#,@RevolAngle@=#360#,@Contrast@=#false#,@Begin Pos@=#517.000#,@Readings A@=#2304#,@Scandirection@=#cr-ca#,@MasterXray@=#On#,@Service@=#On#,@CycleTime@=#0.00#,@ZigZagReconVolume@=#0.00#,@ZigZagScanTime@=#0.00#,@EndPos@=#517#,@SpecialMeas@=#None# |
|---|

Figure 7. Example of an anonymized, raw event text.

As described in Section 2, this event text is further processed into stemmed tokens. We present here the according result for this specific event text. We present in Figure 8 the remaining word stem while punctuation as well as stopwords have been removed. In order to avoid misleading mismatches of values that are rounded differently, we reduce the digits after the decimal point to two.

| 'load', 'scan', 'protocol', 'paty', 'loid', '2.0.123456, 'scan', '1.00', 'scanuid', '1.3.12.2.1107.5.1.4.83004.1234567890, 'scan', 'protocol', 'nam', 'rot00', 'org', 'charact', 'mlorgcharabdom', 'body', 'siz', 'origin', 'mladult', 'scan', 'entry', 'nam', 'rot00', 'kind', 'mlrot', 'entry', 'mod', 'standard', 'autorang', 'cont', 'kv', '120.00', 'mas', '250.00', 'car', 'dos', 'off', 'aec', 'off', 'ctdi', '16.66', 'dlp', '59.98', 'slic', '0.60', 'scan', 'start', 'mlrangestartauto', 'slic', 'wid', 'collim', '60.00', 'no', 'of', 'acquisit', 'slic', '60.00', 'cbc', 'off', 'scan', 'trig', 'mlscantriggerauto', 'no', 'scan', '1.00', 'examin', 'tim', '0.50', 'scantim', '1.00', 'rottim', '0.50', 'rotkind', 'norm', 'currentpeak', '250.00', 'dosemodulationtyp', 'mlnomodulation', 'foc', 'mlsmallfocus', 'anodespee', 'a', '120.00', 'startdelay', '2.00', 'noofclustersperrang', '1.00', 'revolangl', '360.00', 'contrast', 'fals', 'begin', 'pos', '517.00', 'read', 'a', '2304.00', 'scandirect', 'cr-ca', 'masterxray', 'on', 'serv', 'on', 'cycletim', '0.00', 'zigzagreconvolum', '0.00', 'zigzagscantim', '0.00', 'endpo', '517.00', 'specialmea', 'non' |
|---|

Figure 8. The example event text represented as preprocessed tokens.

All further calculations base on the preprocessed tokens. By applying an adapted version of Vaarandi [10], we automatically find the appropriate lines that contain the desired information. Furthermore, we detect clusters in the vast amount of event text. In this example, we learnt a subset of the tokens to be the representative, common subset of all lines. Figure 9 shows the clusters highlighted in the list of preprocessed tokens. Moreover, this cluster represents a subset of the HMM's states which is the basis for all following steps.

```
'load', 'scan', 'protocol', 'paty', 'loid', '2.0.123456, 'scan', '1.00', 'scanuid',
'1.3.12.2.1107.5.1.4.83004.1234567890, 'scan', 'protocol', 'nam', 'rot00', 'org', 'charact',
'mlorgcharabdom', 'body', 'siz', 'origin', 'mladult', 'scan', 'entry', 'nam', 'rot00', 'kind', 'mlrot',
'entry', 'mod', 'standard', 'autorang', 'cont', 'kv', '120.00', 'mas', '250.00', 'car', 'dos', 'off', 'aec',
'off', 'ctdi', '16.66', 'dlp', '59.98', 'slic', '0.60', 'scan', 'start', 'mlrangestartauto', 'slic', 'wid', 'collim',
'60.00', 'no', 'of', 'acquisit', 'slic', '60.00', 'cbc', 'off', 'scan', 'trig', 'mlscantriggerauto', 'no', 'scan',
'1.00', 'examin', 'tim', '0.50', 'scantim', '1.00', 'rottim', '0.50', 'rotkind', 'norm', 'currentpeak',
'250.00', 'dosemodulationtyp', 'mlnomodulation', 'foc', 'mlsmallfocus', 'anodespee', 'a', '120.00',
'startdelay', '2.00', 'noofclustersperrang', '1.00', 'revolangl', '360.00', 'contrast', 'fals', 'begin',
'pos', '517.00', 'read', 'a', '2304.00', 'scandirect', 'cr-ca', 'masterxray', 'on', 'serv', 'on', 'cycletim',
'0.00', 'zigzagreconvolum', '0.00', 'zigzagscantim', '0.00', 'endpo', '517.00', 'specialmea', 'non'
```

Figure 9. The identified cluster is highlighted being a subset of the preprocessed tokens.

The HMM is built based on the preprocessed tokens, found clusters, and emissions. In order to assemble a flexible but precise parser, we search automatically for the token which is followed most often by an emission. In our case, we automatically found "ctdi" emit most values. We integrated here a flexible version as well, as also the trigger for an emission can be subject to changes. For example, from a content point of view, "ctdivol" is an equally good term to denote dose.

In order to parse new data with our found, flexible parser, we phrase the pattern by combining the HMM with the most emitting state. If all tokens in the patterns are found in the event text, the emitted value after "ctdi" is considered as a found emission. The emission found here is 16.660.

### 4.2. Results for Training on Data Set A_train and Validating on Data Set A_test

Following the presented pipeline, we trained our model on $A_{train}$ and then tested it directly without adaption on $A_{test}$. We achieved an accuracy of 99.7% and sensitivity of 52.6%. The respective confusion matrix is given in Table 1.

Table 1. Confusion matrix after training on $A_{train}$ and testing on $A_{test}$.

|  |  | **True Parsing** | |
| --- | --- | --- | --- |
|  |  | Positive | Negative |
| **HMM Parsing** | Positive | 3293 | 0 |
|  | Negative | 2972 | 856393 |

### 4.3. Results for Training on Data Set A and Validating on Data Set B

One of our core targets was to build a flexible parser that is applicable to input from different systems which contain gradually different patterns to be found. Thus, we trained our model on Data A and applied it to Data B. Figure 10 shows an excerpt of the differences in the event texts produced by the both systems, respectively.

| &Load scan protocol&,@Patient LOID@=#2.0.107559#,@Scan@=#1#,@Scan UID@=#1.3.12.2.1107.5.1.4.83004.30000018120111521759000000006#,@Scan protocol name@=#rot00#,@Organ characteristics@=#MlOrgCharAbdomen#,@Body size original@=#MlAdult#,@Scan entry name@=#rot00#,@Kind@=#MlRot#,@Entry Mode@=#standard#,@AutoRange@=#Cont#,@kV@=#120#,@mAs@=#250#,@CARE Dose@=#Off#,@AEC@=#Off#,@CTDI@=#16.660#,@DLP@=#59.975#,@Slice@=#0.6#,@Scan start@=#MlRangeStartAuto#,@Slice Width Collimated@=#60#,@No Of Acquisition Slices@=#60#,@CBC@=#Off# | &Load scan protocol&,@Patient LOID@=#4.0.123727110#,@Scan@=#1#,@StudyLOID@=#1.3.12.2.1107.5.1.4.73307.30000019011600185481300000320#,@Scan protocol name@=#1_HeadSequence#,@Organ characteristics@=#MlOrgCharHead#,@Body size original@=#MlAdult#,@Scan entry name@=#Topogram#,@Kind@=#MlTopo#,@Entry Mode@=#standard#,@AutoRange@=#None#,@kV@=#80#,@mA@=#20#,@CARE Dose@=#Off#,@AEC@=#Off#,@CTDI@=#0.023#,@DLP@=#0.597#,@Slice@=#0.6#,@Scan start@=#MlRangeStartConsole#,@Slice Width Collimated@=#60#,@No Of Acquisition Slices@=#6#,@CBC@=#Off |
|---|---|

Figure 10. Example of an anonymized, raw event text from System A and System B, respectively with the main difference in tokens highlighted.

Without adapting the model, we observe an accuracy of 99.7% and sensitivity of 75.8%. The respective confusion matrix is given in Table 2.

Table 2. Confusion matrix after training on A and testing on B without adaption.

|  |  | **True Parsing** | |
|---|---|---|---|
|  |  | Positive | Negative |
| **HMM Parsing** | Positive | 2113 | 814 |
|  | Negative | 673 | 570663 |

After fitting the model to the new Data B using Baum-Welch and again applying the adapted parsing rules we receive an accuracy of 99.4% and hit rate of 100.0%. The confusion matrix can be found in Table 3, accordingly.

Table 3. Confusion matrix after training on A, fitting model to B and testing on B.

|  |  | **True Parsing** | |
|---|---|---|---|
|  |  | Positive | Negative |
| **HMM Parsing** | Positive | 673 | 3355 |
|  | Negative | 0 | 575558 |

Furthermore, we adapted the model by applying Viterbi to the learnt HMM with the emissions of Data B. This gave us an accuracy of 99.8% and hit rate of 0.0%. We present the respective confusion matrix in Table 4, accordingly.

Table 4. Confusion matrix after training on A, fitting only states to B and testing on B.

|  |  | True Parsing | |
|---|---|---|---|
|  |  | Positive | Negative |
| **HMM Parsing** | Positive | 0 | 4028 |
|  | Negative | 2786 | 578290 |

## 5. DISCUSSION

We built a flexible, learning parser which shall adapt to gradual changes in the machine written input over software versions. In order to evaluate and further improve our approach, we tested our system in different set ups with two different data sources. In the following, we discuss the results in more detail and propose how to interpret those.

### 5.1. Discussion on Intermediate Results along the Pipeline

We evaluated the individual steps of our processing pipeline and showed the results exemplary on one line of event text. Stemming and tokenization are one of the first methods applied. The result presented in Figure 5 shows that preprocessing is necessary, limits the complexity and enables further processing. Based on that finding the most decisive tokens in the entire data set works well as shown in Figure 6. The identified cluster elements primarily describe reoccurring elements and important items of lines which distinguish desired lines from others correctly. However, a few tokens are found to be patterns which are indeed values to be parsed and should not considered as patterns. Otherwise, we would only detect lines which carry the values "off" for a specific token and leave out all other lines with respective value equal to "on". Mistakenly found values to be tokens are taken into consideration because of their high frequency in our texts.

### 5.2. Discussion on Results for Training on Data Set A_train and Validating on Data Set A_test

For further evaluation and applicability of the found clusters to our data we built a statistical model. In order to evaluate the quality and significance of this HMM to represent a flexible, adaptive parser, we trained our model on Data $A_{train}$ and double checked the basic functionality by directly testing with $A_{test}$. This implies that our model does not have to adapt to any changes in the input, yet. We prove our model to be very accurate with 99.7%. As the confusion matrix presented in Table 1 shows, we did not parse any value which is not to be found. However, we received almost as many false negative values as we found true positives which leads to a sensitivity as low as 52.6%. We found this on the one hand to be a success meaning that as soon as we found the crucial line of text, we could parse the appropriate value for sure. On the other hand, we did not find all crucial lines of text. This is a consequence of the previously described scenario where we overfitted our cluster and detected the value "off" to be a cluster candidate. As this is a value not represented by all lines which contain the desired dose value, we miss all the lines where the according value is "on".

### 5.3. Discussion on Results for Training on Data Set A and Validating on Data Set B

Our major goal was to build an adaptive parser which can parse gradually changing inputs without missing relevant information. In order to evaluate our model by this functionality, we trained on data set A and validated on data set B. Without adapting our model to the new data, we already achieve an accuracy of 99.7%. Having a closer look at the confusion matrix presented in Table 2, we see that we miss some true values while others are found to be false positive. As we now trained on the full data set A which led naturally to the correction of overfitting described in the previous section, we found the correct lines to be taken into consideration. Furthermore, during the next step of correctly parsing the exact dose values we did not come across any issue. We encountered an issue during evaluation and calculation of the confusion matrix. As described in the processing step, we round the float values during preprocessing to achieve a constant number of two decimal digits over KPI tables as well as event text. Thus, we did not consider an apparently different way of rounding or cutting which is performed by the existing parser. This leads to parsing the correct values but evaluating them incorrectly as we cannot match them, appropriately.

After we fitted our model to the emissions of data set B, we accomplished a hit rate of one hundred percent. We achieve that as applying Baum-Welch and fitting our model to B implies reducing the restrictiveness of the cluster and its length to only two distinct clusters. This leads to no false negative but 3355 false positives which means that we find a lot more lines and entries which should not have been found.

Finally, Viterbi algorithm was applied to the model. This amplified our clusters to such an extent that we did not parse any correct values. The model got sensitive and more descriptive towards wrong event text lines and patterns and led to a hit rate of zero.

## 6. CONCLUSIONS AND FUTURE WORK

We built a flexible parsing model which is capable of adapting to gradual changes in input structures without missing major output. Our model is adaptive to slight changes in input log file and, thus, parses new input with very high accuracy. By constant learning and fitting our model using Baum-Welch, we continuously adapt our parsing rules to the changes in the input data. However, further analysis and testing is needed for a general conclusion and sensitivity should be improved further. We found that applying Baum-Welch leads to a too vague parsing pattern, where as Viterbi delivers too restrictive rules. Furthermore, our evaluation algorithm contains a limitation due to different precisions in the floating values.

Our results imply that HMMs in combination with other preprocessing methods are very powerful in building new learning models for information retrieval. We extended the state of research around HMMs by applying the statistical model specifically on text data and parsing. Thus, we generated deeper understanding of this field of research's opportunities by adding a new perspective on flexible parsing. Our research contributes to the field of information retrieval as it links parsing, pattern recognition and language processing. Due to our knowledge, we are the first to apply HMMs to machine written text in order to build a flexible, automatically learning parser. In practice, this is a starting point for automatization of information retrieval out of log files for any systems. This leads to data analysts being able to base their algorithms on stable, high quality preprocessed data. Thus, companies installing this model holistically, can reduce their maintenance costs drastically while maintaining business insights throughout their systems lifecycles.

In future work, the model should be enhanced to not only parse one value at a time out of an event text but several tokens delivering values. In our example, an ideal system would parse all values belonging to all tokens from "Patient LOID" to "SpecialMeas". In order to fully replace a manually assembled parser, the model should be enhanced to not only parse values but also detect reoccurring patterns of line pairs. Their relationship should be detected automatically, as well as the determining information. For example, the determining information could be a respective time stamp, based on which the timestamp differences of pairs of lines is then calculated for detecting durations. In further research, even more complex patterns and the appropriate concluding KPI based on more than two lines in the event text should be considered. Furthermore, not only one but more tokens should be taken into account.

# Authors

**Nadine Kuhnert** studied medical engineering to receive her Bachelor's degree in 2013 and finished her Master's in Information and Communication technology in 2015 at the University of Erlangen-Nuremberg.

From 2015 to 2018, she joined Siemens Healthineers and worked on usage data analysis and managed a consultative approach based on market and usage data in order to guide customers to the best fitting options and upgrades for their medical imaging systems. Since 2019, Nadine Kuhnert is product manager of a customer service portal of Siemens Healthineers.

Since 2016, she is working on her PhD on the topic of data wrangling applied to medical imaging system log files at the pattern recognition lab in Erlangen, Germany.

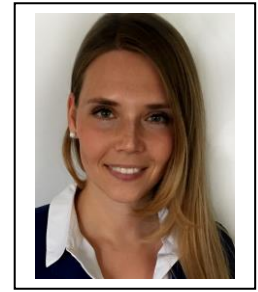

**Prof. Dr. Andreas Maier** studied Computer Science, graduated in 2005, and received his PhD in 2009 at the University of Erlangen-Nuremberg.

From 2009 to 2010, he started working on flat-panel C-arm CT as post-doctoral fellow at the Radiological Sciences Laboratory in the Department of Radiology at the Stanford University. From 2011 to 2012 he joined Siemens Healthcare as innovation project manager and was responsible for reconstruction topics in the Angiography and X-ray business unit.

In 2012, he returned the University of Erlangen-Nuremberg as head of the Medical Reconstruction Group at the Pattern Recognition lab. In 2015 he became professor and head of the Pattern Recognition Lab. Since 2016, he is member of the steering committee of the European Time Machine Consortium. In 2018, he was awarded an ERC Synergy Grant "4D nanoscope". Current research interests focus on medical imaging, image and audio processing, digital humanities, and interpretable machine learning and the use of known operators.

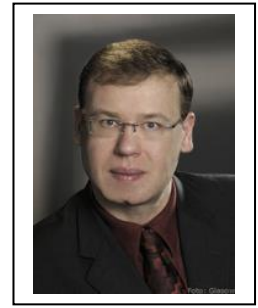